\documentclass{Interspeech}
\usepackage{makecell}

\interspeechcameraready

\title{Multilingual Detection of Alzheimer's Disease from Speech: A Cross-Linguistic Transfer Learning Approach}

\author[affiliation={1}]{Nadine}{Yasser Abdelhalim}
\author[affiliation={1}]{Emmanuel}{Akinrintoyo}
\author[affiliation={1}]{Nicole}{Salomons}

\affiliation{Department of Computing}{Imperial College London}{United Kingdom}
\email{nadine.abdelhalim23@imperial.ac.uk, e.akinrintoyo23@imperial.ac.uk, n.salomons@imperial.ac.uk}
\keywords{speech recognition, human-computer interaction, computational paralinguistics}

\usepackage{comment}

\begin{document}

\maketitle

    
    
    

\begin{abstract}
The development of multilingual Alzheimer's Disease Dementia (AD) detection models presents significant challenges due to the resource-intensive and time-consuming nature of language-specific model training. We propose a novel solution using cross-language training to detect AD in languages beyond those used for model training. This study investigates multilingual deep learning models for detecting AD across different languages and cognitive impairment levels. Using datasets in English, Chinese, Arabic, and Hindi, we developed transformer-based models for binary AD classification. Our approach achieved F1 scores of 82\% across all languages, demonstrating strong cross-linguistic generalization. The rapid inference time (0.5 seconds) supports potential real-time screening applications, while consistent performance across languages indicates feasibility for global deployment.

\end{abstract}

\section{Introduction}
Dementia, a broad term for a range of cognitive impairments, is a progressive neurodegenerative condition that severely impacts memory, reasoning, and daily functioning. Alzheimer's Disease (AD) is the most common form of dementia, affecting an estimated 55 million individuals worldwide, with projections indicating a significant increase to 139 million by 2050 \cite{GALE20181161,chouliaras2023use}. Alarmingly, approximately 75\% of dementia cases remain undiagnosed, which suggests that the actual number of individuals affected is likely far greater than current estimates \cite{gauthier2021world}. While there is no cure for dementia, timely diagnosis is crucial as it allows patients to access treatments that can alleviate symptoms, slow cognitive decline, and improve overall quality of life \cite{rasmussen2019alzheimer}. Early diagnosis and intervention are key to enhancing patient outcomes and providing access to treatments that can mitigate the effects of this debilitating disease.

While advancements in neuro-imaging, such as Magnetic Resonance Imaging (MRI), have improved early AD detection, the high cost of these technologies limits their accessibility in many regions,  particularly in low- and middle-income countries (LMICs) where over 60\% of dementia patients reside \cite{hilabi2023impact,WHO2021dementia}. Concurrently, researchers have increasingly turned to linguistic analysis as a potential tool for detecting early-stage AD. Studies have shown that language impairments, such as speech anomalies and writing difficulties, are early indicators of cognitive decline \cite{mueller2018declines, PAUSES_paper}. Machine learning (ML) models, especially those leveraging linguistic features, have demonstrated success in identifying these early markers of AD \cite{shi_speech_2023}.

However, most of the research in linguistic-based dementia detection has focused primarily on English, creating a significant gap in the ability to detect AD in non-English-speaking populations. With over 7,000 languages spoken worldwide, the reliance on English-centered models overlooks the linguistic diversity of the global population \cite{gauthier2021world}. Moreover, translating existing datasets for every language is an impractical task, particularly for languages with fewer speakers or those with limited research resources. This lack of multilingual datasets hinders the development of universally applicable diagnostic tools for AD, especially in regions where dementia is prevalent but diagnostic tools are scarce.

To address this gap, we propose the development of multilingual models that can detect AD across various languages, without the need for separate datasets for each one. By utilizing multilingual training techniques and cross-lingual learning, these models can generalize across multiple languages, including those with limited research support. This approach has the potential to make early AD detection more accessible and equitable, particularly in LMICs, where the need for such tools is most pressing.

In this research, we detail the development and evaluation of these multilingual models, demonstrating their effectiveness in detecting AD across multiple languages. The findings suggest that multilingual approaches can offer a viable solution to the language barriers that currently limit the reach of AD diagnostic tools, thus enhancing global efforts to identify and manage Alzheimer’s Disease.
\subsection{Linguistic Alzheimer's Disease Detection Models}
The field of Alzheimer's Disease (AD) detection has seen a significant shift towards data-driven approaches in recent years. Szatloczki et al.'s research underscores the efficacy of linguistic analysis of spontaneous speech in detecting AD, surpassing the accuracy of other cognitive examinations. Their work highlights that temporal characteristics of speech, such as tempo, number of pauses, and length, serve as diagnostic markers for early-stage AD, thus motivating the use of linguistic screening in AD diagnosis \cite{szatloczki2015speaking}.

Recent years have witnessed a paradigm shift towards transformer-based architectures, which have consistently outperformed traditional approaches. One of the leading models in dementia detection is a RoBERTa-based model developed by Matoševic and Jovic \cite{ROBERTA1matovsevic2022accurate}. Using a grouped stratified cross-validation approach, their model achieved an F1 score of 90.28\%. The success of this model can be attributed to RoBERTa's ability to automatically extract a wide range of linguistic features, including subtle semantic and syntactic patterns indicative of cognitive decline.

Table \ref{table:Related work} summarises recent Alzheimer's Disease detection research, highlighting the dominance of transformer-based text classification models. 
\begin{table}[htbp!]
\centering
\begin{tabular}{ccc}
Model & F1 & Reference \\ 
\hline
\makecell{RoBERTa}   & 90.28\%  &  \cite{ROBERTA1matovsevic2022accurate} \\
\makecell{ERNIE\\+3Pause} & 88.9\%       & \cite{PAUSES_paper} \\ 
\makecell{BERT\\Large}        & 87.23\%    & \cite{BERTLARGEroshanzamir2021transformer}\\
\makecell{DistilBERT\\+ LR}  & 87\%       & \cite{DistilBertLiu2022transfer} 
\\ 
\makecell{
AWD-LSTM} & 85.19\%       &\cite{AWDLSTMbouazizi2022dementia} 
\\
SVM           & 74\%     & \cite{SVMrimaye2014LearningPL} \\  
\hline
\end{tabular}
\caption{Recent work on AD detection using the Pitt corpus \cite{becker1994natural}, showcasing the models with the highest F1 scores.}
\label{table:Related work}
\end{table}

\subsection{Multilingual and Low-Resource Language Research}
While existing research has demonstrated the efficacy of linguistic analysis for diagnosing AD in well-resourced languages like English, there is a significant gap in diagnostic models for low-resource languages. One primary reason for this disparity is that the majority of publicly available Alzheimer's speech datasets predominantly feature fluent English-speaking participants. 

Few studies have focused on multilingual datasets for Alzheimer's Disease research. 
Pérez-Toro et al. \cite{perez-toro_alzheimers_2022} explored cross-linguistic adaptation between English and Spanish, demonstrating that linguistic features are more critical for AD classification in English, while acoustic features play a larger role in Spanish. By leveraging multilingual embeddings and transfer learning, they demonstrated improved inter-class separability, reinforcing the viability of cross-lingual approaches for AD detection.

For Mandarin, Guo et al. \cite{guo-etal-2020-text} introduced a contrastive learning method and a cross-lingual data augmentation approach to enhance AD detection from speech transcriptions. Their method outperformed conventional CNN-based and BERT-based models, achieving state-of-the-art performance with bilingual data augmentation. The study used an autoencoder to develop shared text representations between Mandarin and English, leveraging a parallel Mandarin-English corpus (OpenSubtitle) to improve classification accuracy.

Rauniyar et al. \cite{rauniyar_breaking_2023} extended this research to Hindi by constructing an AD dataset through manual translation. They prioritized the preservation of cognitive decline markers specific to Hindi grammar and syntax, achieving 85\% agreement between original and translated markers. Their findings emphasize the importance of maintaining linguistic features such as repetitions and grammatical errors that manifest differently across languages.


However, despite these contributions, there is a notable absence of research investigating the effectiveness of multilingual text-based dementia detection models across multiple languages. This gap is significant, as success in generalizing across languages suggests the potential for universally applicable Alzheimer's disease (AD) detection tools, reducing the need for language-specific models. Widely spoken languages like Arabic, Bengali, and Portuguese often lack substantial dementia research datasets, highlighting the importance of this approach. Such multilingual models could be particularly beneficial for countries lacking resources to create extensive language-specific datasets or train separate models.

\section{Datasets}
This section will describe the datasets utilised in this research, detailing their sources, acquisition methods, and relevant characteristics.
\subsection{English Datasets}
The following section outlines the English-language datasets used in this study.
\subsubsection{DementiaBank Pitt Corpus}

\begin{figure}[h!]
\centering
  \includegraphics[width=\columnwidth]{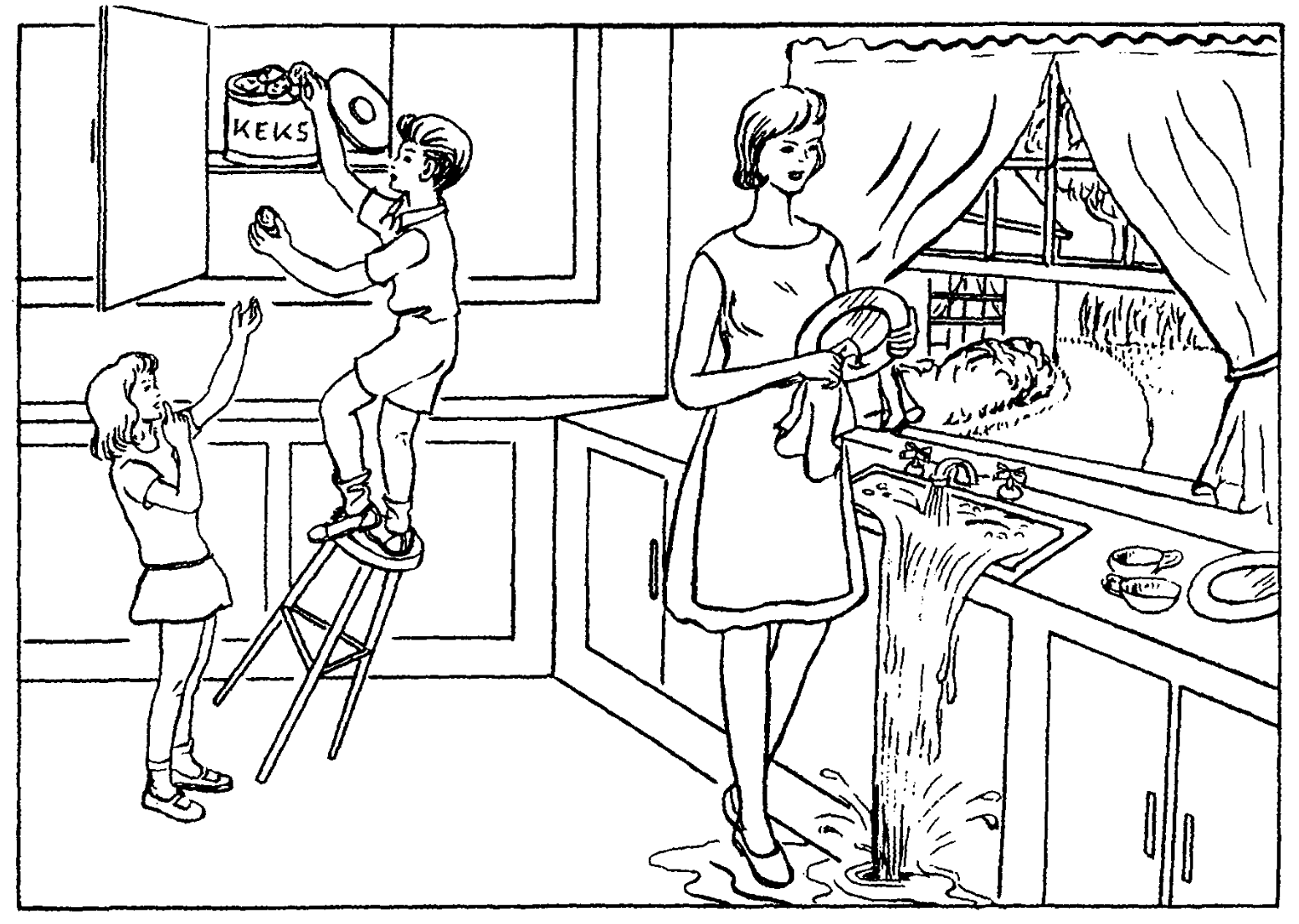}
  \caption{Boston Cookie Theft Picture \cite{cookie_theft}.}
  \label{fig:cookie}
  \vspace{-0.25cm}
\end{figure}

The Pitt Dataset comprises 549 conversation recordings, including 243 from healthy controls and 306 from dementia patients. The participants consisted of elderly controls, people with probable and possible Alzheimer's disease, and people with other diagnoses of dementia. These audio files were collected as part of a larger protocol by the Alzheimer's and Related Dementias Study at the University of Pittsburgh School of Medicine \cite{becker1994natural}. They were then manually transcribed using the CHAT (Codes for the Human Analysis of Transcripts) protocol \cite{macwhinney2000childes,lanzi2023dementiabank}. 


For this research, the analysis specifically focuses on the Cookie Theft picture description task. For the Cookie Theft description task, participants were shown a picture (Figure \ref{fig:cookie}) and instructed to "Describe everything you see going on in this picture". This task elicits spontaneous speech from the participants while providing a standardized context for analysis. The Cookie Theft task is widely used in the study of dementia and is included in the Boston Diagnostic Aphasia Examination (BDAE) \cite{cookie_theft}.

\subsection{Non-English Datasets}

For the purposes of this research, access was granted to three non-English datasets, which will be described in detail in this section.

\subsubsection{DementiaBank Mandarin Lu Corpus}
This dataset consists of audio recordings and corresponding transcriptions from 56 Dementia patients performing the Cookie Theft picture description task in Taiwanese Mandarin, totalling 58 transcriptions. Information on the severity of Dementia for these patients is not included. This dataset is available through DementiaBank \cite{macwhinney2011aphasiabank}. 

\subsubsection{DementiaBankHindi}
The DementiaBankHindi dataset consists of transcripts from 168 patients diagnosed with AD and 98 healthy control participants. This dataset was produced by three fluent Hindi speakers who manually translated the original DementiaBank Pitt Dataset. In addition to the manually translated dataset, they created four additional datasets using neural machine translation \cite{rauniyar_breaking_2023}.

\subsubsection{2024 TAUKADIAL Competition Dataset}
The 2024 TAUKADIAL Competition Dataset \cite{LuzEtAlTAUKADIAL24} consists of Chinese and English speech samples collected from picture description tasks used in cognitive assessments. English-speaking participants aged 60-90 years were recruited in the US, while Chinese-speaking participants aged 60-90 years, with at least six years of education, were recruited in Taiwan. The dataset includes 507 speech samples, with participants classified as either Normal Cognition (NC) or Mild Cognitive Impairment (MCI) based on established criteria.

For the Chinese-speaking participants, picture description tasks included describing a set of three pictures depicting Taiwanese culture.

The transcriptions of the recording were only provided for the English samples. Consequently, following the guidance of the competition organizers, the OpenAI Whisper speech-to-text model. Whisper \cite{whisper} was used to transcribe Chinese samples.

\subsubsection{The Arabic Dataset}
This study also utilized a novel Arabic dataset derived from the English Pitt Dataset. The dataset was created using a systematic translation workflow, combining GPT-4-based machine translation with manual refinement by human annotators to ensure linguistic accuracy and cultural relevance. It comprises 549 transcripts, providing a valuable resource for exploring Alzheimer’s disease detection in Arabic, a low-resource language \cite{Author2025}.
\section{XLM-RoBERTa Model}
The XLM-RoBERTa model is the main model used in this research. XLM-RoBERTa is a multilingual extension of the RoBERTa model, pre-trained on a vast corpus of 100 languages using a masked language modeling (MLM) approach. Unlike earlier multilingual models that relied on the relatively limited scale of Wikipedia data, XLM-RoBERTa was trained on 2.5 TB of cleaned CommonCrawl data, significantly improving its ability to learn cross-lingual representations.
XLM-RoBERTa excels in multilingual tasks such as classification, sequence labeling, and question answering, and demonstrates robust performance on both high-resource and low-resource languages. By setting new performance benchmarks, XLM-RoBERTa has become a key model in multilingual NLP research \cite{conneau-etal-2020-unsupervised}.

\section{Multilingual Models Methodology and Experiments}

In cases where datasets for certain languages do not exist, it is crucial to explore whether AD detection models can transfer across languages. By proving cross-linguistic capabilities, particularly among languages that are not structurally similar, the potential to develop universally applicable diagnostic models increases. 

A series of multilingual models were developed and trained using datasets encompassing a diverse set of languages, specifically English, Chinese, Arabic, and Hindi. The selection of these languages was guided by the practical constraints of data accessibility, as these are the only datasets for which access was granted and that include ground truth labels. 

To achieve this objective, several experiments were carried out involving the fine-tuning and evaluation of multilingual language models on the ‘cookie theft’ task. 
In each experiment, models were trained on datasets from specific languages and tested on an unseen dataset from a different language. The goal was to explore the potential for cross-lingual learning in text-based dementia classification by analyzing how well the multi-lingual classification models performed on languages they had not encountered during training. 

All accessible datasets were utilized, enabling the testing of a variety of languages to determine whether cross-lingual learning was feasible only among linguistically similar languages.
Five main experiments for binary classification were conducted across four languages: English, Arabic, Hindi, and Chinese. 

Each experiment used the XLM-RoBERTa \cite{conneau-etal-2020-unsupervised} architecture and underwent hyperparameter tuning. 
The hyperparameters were tuned across batch sizes of 16 and 32, learning rates from 1e-5 to 5e-5, and epochs ranging from 10 to 45 in 5-step increments, with a maximum sequence length of 512 tokens.

In each of the first four experiments, one language was left out as the test language, while the model was trained on the other three. For example, for the English experiment, the model was fine-tuned on Arabic, Hindi and Chinese and tested on English. This approach aimed to evaluate the model's ability to transfer knowledge from the training languages to an unseen language. The fifth experiment included all languages in both training and testing. The datasets were grouped and organized to prevent the inclusion of the same participants in both training and testing sets, ensuring methodological rigor and avoiding potential bias or overfitting.

The performance of the multilingual models was compared to that of their respective monolingual counterparts, which were fine-tuned solely on the corresponding language datasets. For example, the monolingual English model was fine-tuned solely on the Pitt dataset. Note that a monolingual model for Chinese was not developed as the dataset does not include any speech data from non-dementia patients.

\section{Results and Analysis of the Multilingual Binary Classification Models}
This section presents the outcomes and in-depth analysis of the multilingual classification models for Alzheimer's Disease (AD) detection. The results, summarized in Table \ref{tab:results}, showcase the models' efficacy in different linguistic contexts, highlighting both the challenges and potential of cross-linguistic AD detection.

\begin{table}[htbp]
\centering
\small
\begin{tabular}{@{}clccc@{}}
\toprule
\textbf{Exp} & \textbf{Test Lang.} & \textbf{F1} & \textbf{Epochs} & \textbf{Learning Rate} \\
\midrule
1 & English & 76\% & 30 & 5e-6 \\
2 & Arabic & 71\% & 30 & 5e-6 \\
3 & Hindi & 61\% & 15 & 5e-6 \\
4 & Chinese & 96\% & 30 & 5e-6 \\
5 & All langs & 82\% & 15 & 1e-6 \\
\bottomrule
\end{tabular}
\caption{Results of multilingual binary classification experiments and their optimal hyper-parameters}
\label{tab:results}
\vspace{-0.25cm}
\end{table}

Most experiments favored a learning rate of 5e-6 and 30 epochs, as determined using performance on the development set. However, the Hindi validation set and the comprehensive multilingual model (Experiment 5) achieved optimal results with 15 epochs during tuning. The all-language model further benefited from a lower learning rate of 1e-6, indicating that a more gradual learning process was advantageous when accommodating the dataset's full linguistic diversity. AdamW was identified as the most effective optimizer across all models.

Although the multilingual models underperformed compared to their monolingual counterparts, the results demonstrate the viability of leveraging multilingual datasets implying that it is possible to train multilingual AD detection models for all languages, even those without available dementia datasets. Further expanding the range of languages used during fine-tuning could lead to substantial improvements in model performance. Table \ref{tab:results1} summarizes these comparisons. 

\begin{table}[ht]
    \centering
    \begin{tabular}{|c|c|c|}
        \hline
        \textbf{Language} & \textbf{\makecell{Monolingual \\ Model \\ F1 Score}} 
        & \textbf{\makecell{Multilingual \\ Model \\ F1 Score}} \\
        \hline
        English & 85\% & 76\% \\
        Arabic & 82\% & 71\% \\
        Hindi & 82\% & 61\% \\
        Chinese & - & 96\% \\
        Multilingual & - & 82\% \\
        \hline
    \end{tabular}
    \caption{F1 Scores Comparison Between Monolingual and Multilingual Models (Chinese monolingual model excluded due to absence of non-dementia data).}
    \label{tab:results1}
    \vspace{-0.5cm}
\end{table}

There is a notable performance decline for the Hindi dataset, with the multilingual model's F1 score dropping by 21\% when compared to its monolingual counterpart. While this score surpasses random guessing, the magnitude of the performance drop is significant.


The overall results highlight a robust capability for cross-linguistic generalization in AD detection tasks. Despite slight performance drops, the models' performance remains commendable, considering it had no prior fine-tuning on the test languages. This positive outcome supports the concept of cross-linguistic learning, even in cases where the languages differ substantially and use distinct alphabets.

\section{Inference Time Analysis}
The efficiency of real-time dementia detection applications depends heavily on the model's inference time. To evaluate the system's practical deployment, detailed performance tests were conducted, focusing on inference speed. All tests were performed on an Apple M1 Mac (2020) with an 8-core CPU, 16-core Neural Engine, and a base clock speed of 3.2 GHz.

The classification model's inference time, irrespective of the model variant, consistently hovered around 0.5 seconds, making it feasible to be used in real-time. This uniformity suggests that the choice between different models does not significantly impact the speed of the detection system, thereby affording flexibility in model selection without compromising performance.

When utilizing the OpenAI API for speech-to-text conversion, the Whisper model exhibited an inference time of up to five seconds, with variations depending on the length of the input.
These performance assessments demonstrate the system's capability to effectively handle multilingual input while maintaining responsive interaction.

\section{Discussion}
The significant performance disparity between Hindi and Arabic is particularly intriguing. Despite Hindi being an Indo-Aryan language (part of the Indo-European family, like English), it performed worse than Arabic, which belongs to the entirely separate Semitic language family. This unexpected result suggests that factors beyond broad linguistic relationships are influencing the model's performance. Several hypotheses could explain this finding, including script complexity, syntactic structural differences, and the dataset quality and characteristics. This unexpected result highlights the complexity of cross-linguistic modeling in AD detection.

The strong performance on Chinese (96\% F1 score) and reasonable performance on Arabic (71\% F1 score) and English (76\% F1 score) suggest that these models can capture universal linguistic markers of cognitive decline that transcend specific language families. However, the relatively lower performance on Hindi (61\% F1 score) highlights the complexity of cross-linguistic transfer and the need for further investigation into language-specific features that may impact model performance.

Several promising directions emerge for future work in this field. A primary focus should be the expansion of datasets to include additional languages, particularly from underrepresented language families. This expansion would enhance the model's cross-linguistic generalization capabilities, with special attention given to low-resource languages in regions with limited access to traditional diagnostic tools. Building on this expanded dataset, further investigation into which linguistic features transfer effectively across languages could provide valuable insights for improving model performance and understanding the universal markers of cognitive decline. This analysis could help identify why certain languages, such as Hindi in our study, show lower performance than others, leading to more robust multilingual models.

\section{Conclusion and Future Work} This research demonstrates the feasibility of multilingual AD detection models, achieving an F1 score of 82\% across diverse languages while maintaining rapid inference times of approximately 0.5 seconds. Our findings show that transformer-based models can effectively transfer knowledge across linguistically distinct languages, even when tested on languages absent from the training data. This capability is particularly significant for addressing the global challenge of early AD detection, especially in regions where language-specific datasets and resources are scarce.

Beyond expanding textual data, the integration of multiple modalities presents another promising avenue for research. While our current approach focuses on text analysis, combining this with acoustic features could potentially improve detection accuracy, particularly for languages where the text-only approach shows lower performance. This multimodal approach might better capture the full spectrum of cognitive decline markers across different linguistic contexts.

These advancements could significantly impact global healthcare, particularly in regions where traditional diagnostic resources are limited. By continuing to refine and validate these multilingual approaches, we can work toward more accessible and equitable AD detection tools for diverse populations worldwide.

\bibliographystyle{IEEEtran}
\bibliography{mybib}

\end{document}